\documentclass[11pt]{article}
\usepackage[preprint]{acl}

\usepackage{times}
\usepackage{latexsym}
\usepackage[T1]{fontenc}
\usepackage[utf8]{inputenc}
\usepackage{microtype}
\usepackage{inconsolata}
\usepackage{graphicx}
\usepackage{hyperref}
\usepackage{booktabs}
\usepackage{multirow}
\usepackage{amsmath}
\usepackage{amssymb}
\usepackage{tikz}
\usepackage{pgfplots}
\pgfplotsset{compat=1.18}
\usetikzlibrary{positioning,arrows.meta,fit,backgrounds,calc,shapes.geometric}

\newcommand{\cfg}[1]{\texttt{\small #1}}
\newcommand{\best}[1]{\textbf{#1}}
\newcommand{\ASCR}{\textsc{ascr}}
\newcommand{\ASCRH}{\textsc{ascr-h}}
\newcommand{\DTF}{\textsc{dtf}}
\newcommand{\repourl}{https://github.com/OryCore/Research/tree/master/Annotated\%20Surrogate\%20Retrieval\%20for\%20Polish\%20Statutory\%20Law}

\title{Annotated Surrogate Retrieval for Polish Statutory Law}
\author{Orkun Yiğit Cengiz\thanks{ORCID: \href{https://orcid.org/0009-0003-4299-3629}{0009-0003-4299-3629}} \\
  Independent Researcher \\
\texttt{orkuny.research@gmail.com}}

\begin{document}
\maketitle
\begin{abstract}
  We present a family of retrieval methods for Polish statutory law built on
  \emph{document surrogates}: language-model annotations attached to statutory
  articles at index time. Three designs occupy different points on the
  cost--quality frontier. \ASCR{} is a surrogate cascade with reranking;
  \ASCRH{} fuses a dense list into that cascade; and \DTF{} replaces both
  language-model stages with three lexical and dense retrievers, weighted
  reciprocal rank fusion, and a deterministic re-scoring prior, using no
  model call before generation. We evaluate all three against fourteen
  lexical, dense, fused and ablated baselines plus four controls, on 300
  questions drawn from the 2024 and 2025 Polish bar and legal counsel entrance
  examinations, of which 264 have their reference article in the corpus, over a corpus of 82{,}508 articles from 1{,}133 acts.
  On paired McNemar tests, \ASCRH{} places the reference provision at rank one
  significantly more often than every other non-oracle configuration except one
  of its own ablations (eighteen of twenty comparisons significant in its favour
  at $p<0.005$), reaching 72.3\% against
  61.7\% for BM25 and 52.3\% for dense retrieval. The advantage is concentrated
  at the head and does not survive depth: it is significant at cutoffs of one
  and five, disappears by ten, and by twenty \DTF{} leads on point estimate
  (86.0\% versus 84.5\%) at one ninth the latency and less than half the cost.
  Ablation attributes 27.6 points of rank-one accuracy to the reranking stage
  alone. We further report that the ranking advantage does not extend to
  citation accuracy, where \DTF{} matches the oracle ceiling, and three negative
  results on lemmatisation, pseudo-relevance feedback and query rewriting.
  Surrogate annotation covers 27.0\% of the corpus but every reference provision
  in the benchmark, an asymmetry we disclose and discuss. The benchmark,
  per-question outputs and paired significance tests are publicly available.
\end{abstract}

\section{Introduction}

Identifying the statutory provision that governs a legal question is a
retrieval problem with an unusually narrow target. The answer is a single
article among tens of thousands, and a generator that reads only the head of a
candidate list is not indifferent to where that article was placed. Systems
that appear equivalent at a cutoff of ten can differ by more than twenty points
at rank one.

That the head of the ranking is the quantity of interest is a property of our
ground truth rather than a stylistic preference. Existing statutory article
retrieval datasets label each question with a \emph{set} of relevant
provisions, which makes rank-one measures inapplicable: \citet{louis2023finding} decline to report mean reciprocal rank on BSARD because it would credit only the first relevant article, and, with some questions admitting up to a hundred relevant articles, evaluate instead at recall depths of 100 and beyond. Our
reference labels name a single governing provision in all but six of 300 cases,
so rank-one accuracy is both meaningful and the metric a downstream generator is
most sensitive to.

Polish makes the problem harder. The language inflects nouns and adjectives
across seven cases, so a question and the statute answering it often share
meaning without sharing surface form. Legal benchmarks for the domain
concentrate on English \citep{guha2023legalbench, chalkidis2022lexglue}, or
evaluate classification, named entity recognition and reasoning rather than
retrieval across jurisdictions \citep{niklaus2023lextreme,
multilegalbench2026}. Retrieval-specific legal resources exist for Belgian
\citep{louis2022statutory}, Canadian \citep{canlegalrag2026} and United States
\citep{reasoningfocused2025} law; Polish statutory provisions are not a
retrieval target in any of these three. Polish has been benchmarked for
open-domain passage retrieval \citep{dadas2024pirb}, and \citet{rybak2024silver}
release a Polish dense retriever evaluated on three passage-retrieval datasets,
one of which retrieves over a corpus of 26{,}000 passages extracted from more than 1{,}000
acts of law. \citet{smywinski2025statement} evaluate Polish legal retrieval
across four datasets, one of which pairs legal trainee examination questions
with their cited provisions over a corpus of 3{,}654 provisions assembled from
the answer keys themselves, which its authors describe as posing no real
challenge to current models. We retrieve at article granularity over the full
statutory corpus of 82{,}508 articles, a candidate set 22.6 times larger than
the latter, and report rank-one accuracy rather than Accuracy@10 or nDCG at
depth.

We approach it through \emph{document surrogates}: language-model annotations
generated once per article at index time, consisting of a summary, a theme, a
concept set, and hypothetical questions the article would answer. Retrieval
matches against these derived representations rather than, or alongside, the
verbatim statutory text. We present three designs built on this idea and
evaluate them against fourteen baselines and four controls.

The results separate by depth. \ASCRH{}, which fuses a dense list into the
surrogate cascade before reranking, places the reference provision at rank one
72.3\% of the time, significantly more often than every other non-oracle configuration
tested except one of its own ablations. \DTF{}, which discards both pre-generation model calls in
favour of three-way fusion and a deterministic prior, trails at rank one by
twenty points but catches up by a cutoff of ten and leads on point estimate
from twenty onward, at one ninth the latency. The two systems are complements
rather than competitors: one is built for precision at the head, the other for
coverage at depth under a tight cost budget.

\paragraph{Contributions.}
\begin{enumerate}
    \itemsep0.15em
  \item Three surrogate-based retrieval designs for statutory law, formally
    specified in Section~\ref{sec:methods}, spanning a ninefold range in
    latency.
  \item A controlled evaluation of seventeen configurations and four controls
    under identical corpus, prompt and generation conditions, with paired
    significance tests on 300 examination questions, reported at six cutoffs.
  \item A depth-dependent convergence: reranking dominates at rank one, the
    two designs can no longer be separated at a cutoff of ten on this sample,
    and deterministic fusion leads on point estimate from twenty
    (Figure~\ref{fig:crossover}).
  \item An ablation isolating each component, and three negative results:
    lemmatisation, pseudo-relevance feedback and query rewriting each fail
    to improve retrieval.
\end{enumerate}

\section{Related Work}\label{sec:related}

\paragraph{Statutory article retrieval.}
The task of returning the law articles relevant to a legal question was
formalised for a non-English jurisdiction by \citet{louis2022statutory}, whose
Belgian dataset (BSARD) pairs 1{,}108 expert-labelled French questions with
articles drawn from a corpus of 22{,}633; their strongest \emph{supervised}
baseline, a trained bi-encoder, reaches 74.8\% Recall@100 on the 222-question
test split, while the best zero-shot baseline reaches 51.3\%.
\citet{louis2023finding} augment a dense retriever with the hierarchical structure of the legislation. Our setting is the Polish analogue at 3.6 times the corpus size, with reference
provisions supplied by a state examination commission rather than by annotators
hired for the purpose. BSARD questions are multi-label, with up to a hundred
relevant articles each; ours name a single governing provision in all but six of
300 cases, which is what makes rank-one accuracy available as a metric here and
not there
\citep{louis2023finding}.

\citet{reasoningfocused2025} come closest to our query type but not to our
target: their Bar Exam QA task poses examination hypotheticals against a pool of gold
explanation passages, paragraph-level United States caselaw and encyclopedia
entries, while their Housing Statute QA task targets statutes but with binary
questions. Both are reported as challenging for baseline retrievers.
Retrieving a statutory article in response to an examination question falls
between the two.

\paragraph{Legal retrieval benchmarks.}
LegalBench-RAG \citep{pipitone2024legalbenchrag} established retrieval
evaluation over legal corpora as distinct from end-to-end question answering.
Later work extends this to legal retrieval requiring reasoning
\citep{reasoningfocused2025}, clause-level grounding in
{Chinese} legal documents \citep{legaldc2026}, underrepresented jurisdictions
\citep{canlegalrag2026}, and end-to-end pipeline evaluation
\citep{legalragbench2026}. Broader legal benchmarks
\citep{guha2023legalbench, chalkidis2022lexglue, niklaus2023lextreme} measure
classification, named entity recognition and reasoning rather than retrieval
quality. \citet{multilegalbench2026} does include Polish, and LEXTREME
\citep{niklaus2023lextreme} covers Polish within several of its multi-jurisdiction
datasets, but in every case the tasks are classification, named entity recognition and norm extraction rather than retrieval, and no statutory corpus is indexed for search. Polish national \emph{legislation} is not a retrieval
target in any of the benchmarks listed above.

\paragraph{Polish retrieval.}
PIRB \citep{dadas2024pirb} evaluates dense and hybrid retrieval across 41
Polish tasks and identifies \texttt{multilingual-e5-large}
\citep{wang2024multilingual} as the prior state of the art among general-purpose
multilingual encoders, which motivates our choice;
\citet{rybak2024silver} release a Polish-specific dense retriever together with
five new passage-retrieval datasets, and evaluate the model on three
passage-retrieval datasets. One of the three, Legal Questions, is a Polish legal retrieval task: 718 questions over
a corpus of 26{,}000 passages extracted from more than 1{,}000 acts of law. It is the closest existing resource to ours in corpus and task form, and we differ from it on four
axes as a retrieval task. Our corpus is 82{,}508 articles rather than 26{,}000 passages; our
retrieval unit is the statutory article rather than a passage; our reference
labels are supplied by a state examination commission rather than by a shared
task; and we report rank-one accuracy rather than Accuracy@10 and nDCG@10. PIRB's law-domain web
subsets, \cfg{specprawnik} and \cfg{e-prawnik}, are built from questions
answered by lawyers rather than from statutes, so the retrieval target there is
an answer text and not an article of law.

\citet{dadas2024ranking} find that most rerankers generalise poorly for Polish
and underperform a strong dense retriever, with the explicit exception of
models with a large parameter count; their remedy is domain fine-tuning or a
larger reranker. Our reranker is a small general-purpose language model rather than one of the
dedicated Polish rerankers they evaluate, so their finding would predict little
benefit from it. Our reranking ablation (Section~\ref{sec:ablations}) shows the
opposite, and we discuss why.

\citet{smywinski2025statement} evaluate Polish legal retrieval across four
datasets. Their LQuAD-PL dataset draws on the same source as ours, legal
trainee entrance examinations, but retrieves over a corpus of 3{,}654
provisions taken from the answer keys; the authors report that current encoders
reach almost perfect nDCG@5 on it and conclude that it poses no real challenge.
They further caution that when questions are derived directly from the target
documents, measured performance is optimistic. We adopt both observations: we
retrieve over the full 82{,}508-article corpus rather than the answer set, and we report the lexical overlap of examination questions as a limitation (see Limitations).

\paragraph{Retrieval and fusion.}
We build on standard lexical \citep{robertson2009bm25} and dense
\citep{karpukhin2020dpr} retrieval, evaluated in the zero-shot regime that
\citet{thakur2021beir} identify as the realistic one for new domains.
Reciprocal rank fusion \citep{cormack2009rrf} combines ranked lists without
requiring score comparability and remains a strong baseline. Cross-encoder
reranking \citep{nogueira2019passage} and language models used directly as
rerankers \citep{sun2023chatgpt, ma2024finetuning} improve head precision at
substantial inference cost, quantified for listwise language-model reranking by
\citet{sun2023chatgpt}.

\paragraph{Generated representations.}
Retrieval-augmented generation \citep{lewis2020rag} makes the quality of the
retrieved set directly consequential for the generated answer. Two established
responses to vocabulary mismatch sit on opposite sides of the index:
\citet{nogueira2019doc2query} expand documents at index time with predicted
queries, while \citet{gao2023hyde} generate a hypothetical document from the
query at retrieval time. Pseudo-relevance feedback
\citep{rocchio1971, lavrenko2001relevance} likewise expands the query.
\citet{nogueira2019doc2query} already cross the two axes factorially for
open-domain passage retrieval; our contribution is to run the same separation
for statutory retrieval, where the document side is a legal norm rather than a
web passage, and to report where the two representations complement rather
than replace verbatim text.

\paragraph{Evaluating legal question answering.}
\citet{karp2026judge} evaluate language models on a Polish professional legal
examination and find automatic scoring by a language model unreliable for the
free-text written component, where model-assigned scores diverged sharply from
those of the examining committee. Their multiple-choice component, by contrast,
separates configurations cleanly. \citet{reasoningfocused2025} report the
complementary result for retrieval: gains in Recall@10 do not reliably
translate into downstream question-answering gains, because improvement is
bounded by how much the reader can extract from the gold passage at all. We
observe the same bound on our task, in a sharper form: answer accuracy for our
generator on this examination is saturated and cannot evaluate retrieval
(Section~\ref{sec:metrics}).

\section{Benchmark and Metrics}\label{sec:benchmark}

\subsection{Source}

The Polish Ministry of Justice publishes, after each annual entrance
examination for the legal traineeships, the complete question set with an
official answer key. We use the joint advocate and legal counsel examination
(\emph{egzamin wstępny na aplikację adwokacką i radcowską}) for 2024 and 2025:
150 questions each, 300 in total. Each is single-choice with three options and
names its governing act in the stem, in the form \emph{``Zgodnie z Kodeksem
karnym, \ldots''}. Answer keys record the correct option and its legal basis as
an article reference such as \emph{art.\ 11 § 3 k.k.}, giving article-level
ground truth produced by a state examination commission.

Two exclusions apply. Nine of the 300 items carry a legal basis our parser
could not resolve to any article, and a further 27 resolve to articles that are
not present in our corpus. The remaining 264 questions, 133 from 2024 and 131
from 2025, form the \emph{retrievable subset}
$n^{*}$.\footnote{Parsed questions, per-question outputs for every
configuration, and all paired significance tests: \url{\repourl}.}

\begin{table*}[t]
  \centering\small
  \begin{tabular}{@{}lr@{\hspace{5mm}}lr@{\hspace{5mm}}lr@{}}
    \toprule
    \multicolumn{2}{@{}l}{\textbf{Benchmark}} &
    \multicolumn{2}{l}{\textbf{Corpus}} &
    \multicolumn{2}{l@{}}{\textbf{Protocol}}\\
    \cmidrule(r){1-2}\cmidrule(lr){3-4}\cmidrule(l){5-6}
    Questions $n$              & 300      & Acts                  & 1{,}133   & Configurations        & 17 + 4 ctrl \\
    Examination years          & 2024--25 & Articles              & 82{,}508  & Retrieval depth $D$   & 100 \\
    Options per question       & 3        & Chunks (embedded)     & 159{,}434 & Context size $k$      & 10 \\
    Reference resolved         & 291      & With surrogates       & 22{,}241  & Generator             & \cfg{flash-lite} \\
    Retrievable $n^{*}$        & 264      & Coverage, corpus      & 27.0\%    & Reranker              & \cfg{flash-lite} \\
    Strata (2024 / 2025)       & 133 / 131& Coverage, reference   & 100\%     & Encoder               & \cfg{mE5-large} \\
    Ref.\ with § / \emph{ust.} & 69\%     & Mean article length   & 1{,}334 ch& Temperature           & 0 \\
    Act named in stem          & 100\%    & Surrogate generator   & \cfg{flash-lite} & Runs per config & 1 \\
    \bottomrule
  \end{tabular}
  \caption{Benchmark, corpus and protocol. Retrieval is the only variable across
    configurations: corpus, prompt, generator, temperature and context size are held
    fixed. \cfg{flash-lite} is \texttt{gemini-3.1-flash-lite} and \cfg{mE5-large} is
    \texttt{multilingual-e5-large}; the same model generates the surrogates, performs
    query analysis, reranks and generates the answer. ``Coverage, reference'' is the
    fraction of the 264 retrievable reference articles carrying a surrogate
    annotation; see Limitations. The generator context is
    additionally capped at a 6{,}000-token budget, and individual articles at
  6{,}000 characters, so a retrieved article may be truncated or dropped.}
  \label{tab:setup}
\end{table*}

\subsection{Corpus and surrogates}

The corpus comprises 1{,}133 acts segmented into 82{,}508 articles and
159{,}434 embedded chunks, an average of 1.9 chunks per article. Because
reference labels are article-level, chunk scores are max-pooled to their
containing article, $s(a) = \max_{c \in a} s(c)$. Max-pooling gives longer
articles more opportunities to score; at this chunk-to-article ratio the effect
is small, but it is not zero and we do not correct for it.

A subset of 22{,}241 articles carries a language-model annotation: a summary
$\sigma_a$, a theme $\theta_a$, a concept set $C_a$, and a set of hypothetical
questions $Q_a$ the article would answer. Following standard information
retrieval terminology we call these a \emph{document surrogate}, a derived
representation standing in for the document during matching.

\paragraph{Coverage is asymmetric, and the asymmetry is not incidental.}
Surrogate coverage is 27.0\% of the corpus but 100\% of the reference
provisions in the benchmark. Annotation was prioritised by act importance,
chiefly the major codes and the Constitution, and examination questions draw
disproportionately on those same acts, so the two selections are correlated
rather than independent. Configurations that match against surrogate fields
therefore score a fully annotated target against a candidate field that is
largely unannotated. This applies to \ASCR{}, \ASCRH{} and its four ablations,
\cfg{bm25-surrogate}, \cfg{dense-surro-rrf}, \cfg{dense-surro-rescore},
\cfg{dense-prf}, and the $R_{\mathrm{cov}}$ branch of \DTF{}. Some unknown share of the rank-one
advantage those configurations show is attributable to this asymmetry rather
than to the retrieval mechanism, and we do not have an experiment that
separates the two. The Limitations section states what this does and does
not threaten.

The retrieval corpus is the full statutory collection rather than the set of
provisions appearing as examination answers. The closest prior resource in question source,
LQuAD-PL \citep{smywinski2025statement}, draws on the same examinations but
retrieves over 3{,}654 provisions assembled from the answer keys, a candidate
set 22.6 times smaller. Our absolute figures therefore sit below theirs, on a task that is correspondingly harder.

\subsection{Metrics}\label{sec:metrics}

We report Hit@$k$ for $k \in \{1,5,10,20,50,100\}$, the fraction of questions for
which a reference article appears in the top $k$, together with MRR and
nDCG@10, all over the retrievable subset $n^{*}=264$ and marked with an
asterisk. On this subset the oracle control attains Hit@$k$ of 100\% by
construction. Every starred percentage in the paper corresponds to an integer
count out of 264. Point estimates carry 95\% Wilson score intervals
\citep{wilson1927}. Citation accuracy and answer accuracy are reported over all
$n=300$ questions, since a question whose reference article is missing from the
corpus still has a correct option and a correct legal basis, either of which the
generator may supply from parametric knowledge; the closed-book control names
the reference provision for 141 of 300 questions with no retrieval at all.

A small number of questions carry more than one reference article. Hit@$k$
counts such a question as correct when any reference article appears in the top
$k$, and MRR and nDCG@10 likewise use the highest-ranked reference article. We
do not report a fractional recall, which would penalise the oracle control for
injecting one article rather than all of them. Six of the 300 questions are
affected: three in each examination year, five naming two articles and one
naming three.

\paragraph{Citation accuracy} measures whether the model named the reference
provision when asked for the legal basis of its answer. Only the first cited
article is scored, so verbosity confers no advantage; act matching is
boundary-anchored so that \cfg{k.p.} does not match inside \cfg{k.p.k.}; and
unparseable citations are scored incorrect. Matching is article-level, and 69\%
of reference provisions additionally specify a subdivision that is not scored.
We also report the fraction of the achievable band recovered,
\begin{equation}
  \label{eq:gap}
  G(m)=\frac{\mathrm{cit}(m)-\mathrm{cit}(\textsc{closed-book})}
  {\mathrm{cit}(\textsc{oracle})-\mathrm{cit}(\textsc{closed-book})}
\end{equation}
with floor $141/300=47.0$ and ceiling $211/300=70.3$, an achievable band of
23.3 points.

\paragraph{Answer accuracy is not a retrieval metric here.}
A closed-book control receiving no retrieved context answers 92.7\% (278/300)
of questions correctly, against 96.3\% (289/300) for \ASCRH{} and 94.3\%
(283/300) for the oracle control, which receives the reference article directly;
neither \ASCRH{} against \cfg{closed-book} ($b{=}21$, $c{=}10$, $p=0.072$) nor
\ASCRH{} against the oracle ($b{=}14$, $c{=}8$, $p=0.29$) is significant. Every retrieval configuration falls between 89.3\% and 97.7\%.
That the oracle answers slightly worse than \ASCRH{} despite receiving the
reference article is within noise at this sample size, but it is a further sign
that the generator is not relying on the retrieved context to choose an
option. A three-option question is
answerable by elimination, and our generator, which has Polish legal material in
its training distribution, eliminates effectively without consulting a source.
We report answer accuracy for completeness only.

This is a property of our generator and this examination rather than of the
format in general. \citet{karp2026judge} report closed-book accuracies in the mid-60s to high-70s on a Polish three-option legal knowledge test, with substantial gains from retrieval, so the same format does separate configurations when the generator is not saturated. Nor should the argument be generalised across option counts: the
Bar Exam QA task of \citet{reasoningfocused2025} uses four-option items, for
which elimination is correspondingly harder.

\paragraph{Significance.}
Comparisons use McNemar's test \citep{mcnemar1947} with the continuity
correction of \citet{edwards1948} on per-question binary outcomes. The two
examination years are independent strata; we compute discordant cells within
each year, sum them across strata, and test the pooled counts, reporting the
discordant totals alongside each $p$-value, since these and not the difference
in point estimates determine the power of the test. We apply no
multiple-comparison correction, but note that Holm--Bonferroni over the twenty
rank-one comparisons of Table~\ref{tab:mcnemar} leaves every conclusion
unchanged: the two largest significant $p$-values, both $0.004$, remain below
$0.05$ after adjustment.

\section{Methods}\label{sec:methods}

\paragraph{Notation.}
$q$ is a question, $a$ an article, $\xi(a)$ its verbatim text, $d(a)$ its act,
$\mathcal{D}$ the set of acts, $E(\cdot)$ the mE5 encoder, and
$\sigma_a,\theta_a,C_a,Q_a$ the surrogate fields of Section~\ref{sec:benchmark}.
$\tau(q)$ denotes the set of content terms of $q$ after stop-word removal, and
$\tilde q$ the reformulated question produced by the query-analysis call of
\ASCR{}. $h(Y,\tau(q))$ is the term overlap
between a field $Y$ and $\tau(q)$; $r(Y,y)$ is a normalised text-similarity
score in $[0,1]$, abbreviated $r(Y)$ when the second argument is the raw
question. All configurations return up to $D{=}100$ articles; the generator receives at
most $k{=}10$ of them, subject to a 6{,}000-token context budget.

\subsection{Shared primitives}

Lexical matching uses BM25 \citep{robertson2009bm25} with inverse document
frequency, term saturation and length normalisation, over an indexed text $x$:
\begin{equation}
  \mathrm{BM25}(q,x)=\sum_{t \in q}\mathrm{idf}(t)\,
  \frac{f_{t,x}(k_1{+}1)}{f_{t,x}+k_1\!\left(1-b+b\frac{|x|}{\overline{|x|}}\right)}
\end{equation}
with $k_1{=}1.2$, $b{=}0.75$, the common defaults;
\citet{robertson2009bm25} note that the model itself gives no guidance on these
values and that optima are collection-dependent, and we did not tune them
(see Limitations).

Rank fusion uses a weighted variant of reciprocal rank fusion
\citep{cormack2009rrf}, which is insensitive to incomparable score scales:
\begin{equation}
  \label{eq:rrf}
  s_{\mathrm{RRF}}(a)=\sum_i \frac{w_i}{k_0+\mathrm{rank}_{R_i}(a)},
  \qquad k_0=20
\end{equation}
The original formulation carries no weights and fixes $k=60$, a value its
authors report as near-optimal but not critical in their own pilot sweep. Our $k_0{=}20$ and the weights $w_i$ were set by hand for this task and
not tuned on a development split (see Limitations); neither value follows from their results.

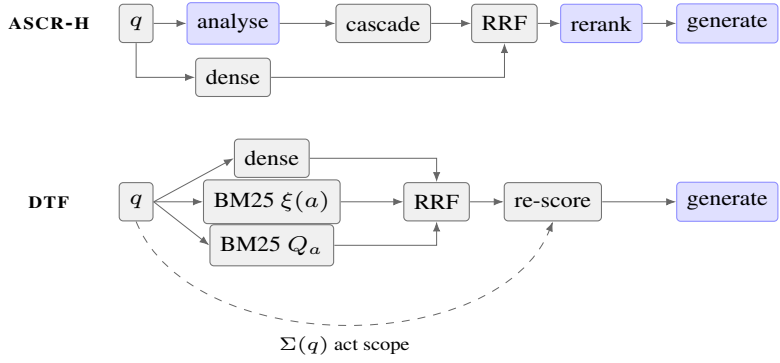
\begin{figure*}[t]
  \centering
  \begin{tikzpicture}[xscale=1.28, yscale=1.15, transform shape,
      font=\scriptsize,
      box/.style={draw=black!55, rounded corners=1.5pt, minimum height=4.4mm,
      inner xsep=3pt, align=center},
      llm/.style={box, fill=blue!12, draw=blue!45},
      det/.style={box, fill=black!6},
    ar/.style={-{Latex[length=1.3mm]}, draw=black!60, line width=0.35pt}]

    \node[font=\scriptsize\bfseries] at (-0.15,0) {\ASCRH{}};
    \node[det] (q1)  at (0.75,0)  {$q$};
    \node[llm] (an)  at (1.75,0)  {analyse};
    \node[det] (cas) at (3.30,0)  {cascade};
    \node[det] (dn)  at (1.75,-0.62) {dense};
    \node[det] (rf1) at (4.55,0)  {RRF};
    \node[llm] (rr)  at (5.55,0)  {rerank};
    \node[llm] (g1)  at (6.85,0)  {generate};

    \draw[ar] (q1)  -- (an);
    \draw[ar] (an)  -- (cas);
    \draw[ar] (cas) -- (rf1);
    \draw[ar] (q1.south) |- (dn.west);
    \draw[ar] (dn.east) -| (rf1.south);
    \draw[ar] (rf1) -- (rr);
    \draw[ar] (rr)  -- (g1);

    \node[font=\scriptsize\bfseries] at (-0.15,-2.05) {\DTF{}};
    \node[det] (q2) at (0.75,-2.05) {$q$};
    \node[det] (t1) at (2.15,-1.55) {dense};
    \node[det] (t2) at (2.15,-2.05) {BM25 $\xi(a)$};
    \node[det] (t3) at (2.15,-2.55) {BM25 $Q_a$};
    \node[det] (rf2) at (3.85,-2.05) {RRF};
    \node[det] (bo) at (5.05,-2.05) {re-score};
    \node[llm] (g2) at (6.85,-2.05) {generate};

    \draw[ar] (q2.east) -- (t1.west);
    \draw[ar] (q2.east) -- (t2.west);
    \draw[ar] (q2.east) -- (t3.west);
    \draw[ar] (t1.east) -| (rf2.north);
    \draw[ar] (t2.east) -- (rf2.west);
    \draw[ar] (t3.east) -| (rf2.south);
    \draw[ar] (rf2) -- (bo);
    \draw[ar] (bo)  -- (g2);
    \draw[ar, dashed] (q2.south) to[out=-70,in=-110]
    node[below, font=\tiny, pos=0.5] {$\Sigma(q)$ act scope} (bo.south);
  \end{tikzpicture}
  \caption{The two extremes of the family. \ASCRH{} issues three model calls
    (shaded); the dense branch runs concurrently with query analysis and so adds no
    wall-clock time. \DTF{} issues one, replacing analysis and reranking with
    three-way fusion and a deterministic re-scoring prior that exploits the act
  named in the question stem.}
  \label{fig:arch}
\end{figure*}

\subsection{\ASCR{}: surrogate cascade}

\ASCR{} retrieves \emph{against} the surrogate annotation in two stages. A document stage scores acts and retains the top forty acts,
\begin{equation}
  \begin{split}
    s_{\mathrm{doc}}(d) &= 3h(C_d,\tau(q)) + 2r(\mathrm{title}_d) \\
    &\quad + 2r(\sigma_d) + \mathbb{1}[\phi_d{=}f_q]
  \end{split}
\end{equation}
Here $C_d$ and $\sigma_d$ are act-level fields, formed by pooling the concept
sets and summaries of the annotated articles of $d$; $\phi_d$ is the legal
domain label of act $d$ and $f_q$ the domain the query-analysis call assigns to
$q$, so the indicator rewards a domain match. The stage additionally applies a
bonus of $10$ to acts in a designated tier (\cfg{tier1Bonus} in the released
configuration). Its coefficient is more than three times the weight on
any other term in the stage, so where it differentiates acts it is the dominant
signal.

An article stage scores within those acts,
\begin{equation}
  \begin{split}
    s_{\mathrm{art}}(a) &= 3h(\theta_a,\tau(q)) + 2h(C_a,\tau(q)) + 20\,r(Q_a,\tilde q) \\
    &\quad + 2r(\sigma_a) + 1000\,\mathbb{1}[a \in R_q]
  \end{split}
\end{equation}
where $R_q$ is the set of articles explicitly referenced in the question text.
The coefficient $1000$ is not a weight but an override: an article named in the
question is promoted above all others regardless of the remaining terms. A
language model reranker then reorders the forty highest-scoring articles in a single
listwise call, rather than working in a sliding window; this is what keeps
\ASCRH{} to three model calls in total, including generation.

\subsection{\ASCRH{}: cascade with fused dense evidence}

\ASCRH{} adds a dense candidate list, fused by Eq.~\ref{eq:rrf} before
reranking. The dense branch consumes the raw question and therefore runs
concurrently with the query-analysis call, adding no wall-clock time. Its
effect is to expose the reranker to articles in acts the document stage
discarded entirely, which is where the cascade alone loses recall.

\subsection{\DTF{}: deterministic tri-signal fusion}

\DTF{} targets the same geometry with cheaper components. Three retrievers with
complementary failure modes are fused and then re-scored without any model call:
\begin{align}
  R_{\mathrm{den}}&=\operatorname*{arg\,top}\nolimits^{D}\max_{c\in a}\cos(E(q),E(c))\\
  R_{\mathrm{txt}}&=\operatorname*{arg\,top}\nolimits^{D}\mathrm{BM25}(q,\xi(a))\\
  R_{\mathrm{cov}}&=\operatorname*{arg\,top}\nolimits^{D}\mathrm{BM25}(q,Q_a)
\end{align}
Dense retrieval recovers paraphrase and inflection but misses precise statutory
terminology; matching over $\xi(a)$ captures verbatim quotation but not
paraphrase; matching over $Q_a$ closes the gap between how questions are asked
and how statutes are written. Fusion applies Eq.~\ref{eq:rrf} with
$(w_{\mathrm{den}},w_{\mathrm{txt}},w_{\mathrm{cov}})=(1.0,0.7,1.0)$, followed
by a deterministic re-score:
\begin{multline}
  \label{eq:rescore}
  s(a)=s_{\mathrm{RRF}}(a)+\beta\,\mathbb{1}[d(a)\in\Sigma(q)]
  +\tau_0\,\mathbb{1}[\theta_a\cap q]\\
  +\gamma\min(3,|C_a\cap q|)-\rho\,\mathbb{1}[\text{repealed}]
  +\pi\,\mathbb{1}[a\in R_q]
\end{multline}
with $(\beta,\tau_0,\gamma,\rho,\pi)=(0.35,0.04,0.02,0.15,1.0)$.

\paragraph{The coefficients are not all on the same scale, deliberately.}
With $k_0{=}20$ and these weights, an article ranked first in all three lists
attains $s_{\mathrm{RRF}}=2.7/21=0.129$, so the fused score occupies
$[0,0.129]$. $\tau_0$ and $\gamma$ are genuine priors that break ties within
the fusion scale, and with them the maximum attainable without $\beta$ or $\pi$
is $0.229$. Against that ceiling $\rho{=}0.15$ is a strong but not decisive
penalty. $\beta$ and $\pi$ are decisive: any non-repealed article in an act
matched by $\Sigma(q)$ outranks every article outside it whatever the retrieval
evidence, and an article named in the question outranks everything. The
exception is instructive: a repealed in-scope article floors at roughly
$0.208$, just below the $0.229$ available out of scope, so the repeal penalty
can pull an article back across the scope boundary. We state this
plainly because it determines how the results should be read: \DTF{}'s
behaviour on this benchmark is that of act-scoped fusion, not of fusion with a
soft preference for the scoped act. The additive form rather than an explicit
filter still matters: the ordering \emph{within} and \emph{outside} the scoped
act remains the fused ordering, and articles outside the scope are demoted
rather than deleted, so recall at depth would survive a wrong scope match. That
robustness is untested here, since the matcher identified the correct act on
every question in the benchmark. The head of the list is determined by
$\Sigma(q)$.

\paragraph{Act scoping.}
Every examination item names its governing act in the stem, a signal every
other configuration discards. We extract the act phrase from the stem,
normalise its tokens to five-character prefixes, which collapses Polish case
endings without a lemmatiser (\emph{Kodeksem karnym} $\rightarrow
\{$\emph{kodek}, \emph{karny}$\}$), and match by containment against the
prefix sets of act titles. Writing $P(q)$ for the prefix set of the extracted
act phrase and $P(d)$ for that of the title of act $d$,
\begin{equation}
  \Sigma(q)=\left\{d\in\mathcal{D}:\frac{|P(q)\cap P(d)|}{|P(q)|}\ge 0.6\right\}
\end{equation}
$P(q)$ covers only the act phrase, not the whole question; the
threshold is not attainable against a full question's token set.

Note that $\beta$ re-scores candidates already returned by the three retrievers
rather than retrieving from $\Sigma(q)$ directly, so a scoped act with no
article in any of the three lists receives no benefit. $\Sigma(q)$ is also a set
rather than a single act, and where containment admits more than one title
each matching act receives the full boost; we do not report how many acts it typically contains. Either mechanism
would account for \DTF{}'s document-level hit rate of 92.0\% rather than 100\%
despite a scope match on every question, and we do not separate them.

\subsection{Baselines and controls}

Lexical: \cfg{bm25-raw} over $\xi(a)$; \cfg{bm25-lemma} with suffix stripping;
\cfg{bm25-surrogate} over $\sigma_a\Vert\theta_a\Vert C_a\Vert Q_a$;
\cfg{bm25-expanded} with a model-expanded query. The last three separate
index-side from query-side generation. Note that \cfg{bm25-surrogate}
\emph{substitutes} the surrogate fields for the statutory text, whereas
\citet{nogueira2019doc2query} append predicted queries to the document and
retain it; the corresponding concatenated index, BM25 over
$\xi(a)\Vert Q_a$, is not among our configurations, and
the Limitations section notes what that leaves untested.
Dense: \cfg{dense}; \cfg{dense-rephrased} with a rewritten query.
Fused: \cfg{rrf} (BM25 + dense); \cfg{dense-surro-rrf};
\cfg{dense-surro-rescore}; and \cfg{dense-prf}, a pseudo-relevance feedback
variant whose expansion vocabulary is authored at index time.
Ablations remove one component from \ASCRH{}: \cfg{no-rerank},
\cfg{no-analysis}, \cfg{no-covers}, \cfg{no-concepts}.
Controls: \cfg{closed-book} (no retrieval), \cfg{random}, \cfg{oracle}
(reference article injected), and \cfg{oracle-doc} (correct act, random
articles within it). This gives seventeen configurations and four controls.

\section{Results}\label{sec:results}

\begin{table*}[t]
  \centering\small
  \begin{tabular}{lrrrrrrrrrr}
    \toprule
    & \multicolumn{6}{c}{\textbf{Hit@$k$*}} & \multicolumn{2}{c}{\textbf{Ranking}} & \multicolumn{2}{c}{\textbf{Citation}}\\
    \cmidrule(lr){2-7}\cmidrule(lr){8-9}\cmidrule(lr){10-11}
    Configuration & 1 & 5 & 10 & 20 & 50 & 100 & MRR* & nDCG* & Acc. & $G$\\
    \midrule
    \ASCRH{}                  & \best{72.3} & \best{81.1} & \best{83.7} & 84.5 & 86.0 & 87.5 & \best{0.764} & \best{0.776} & 67.3 & 0.87\\
    \cfg{no-concepts}         & 68.2 & 74.2 & 75.0 & 75.8 & 77.7 & 83.3 & 0.710 & 0.713 & 59.7 & 0.54\\
    \ASCR{}                   & 67.0 & 76.1 & 77.3 & 78.0 & 80.3 & 83.0 & 0.713 & 0.721 & 61.3 & 0.61\\
    \cfg{no-analysis}         & 64.8 & 72.7 & 73.9 & 74.2 & 75.8 & 79.9 & 0.684 & 0.694 & 59.3 & 0.53\\
    \cfg{no-covers}           & 62.9 & 72.0 & 73.5 & 73.5 & 75.4 & 79.5 & 0.670 & 0.679 & 57.0 & 0.43\\
    \cfg{rrf}                 & 61.7 & 75.8 & 80.3 & 83.3 & 85.6 & 86.7 & 0.686 & 0.708 & 67.7 & 0.89\\
    \cfg{bm25-raw}            & 61.7 & 73.9 & 76.1 & 79.2 & 83.0 & 84.1 & 0.677 & 0.690 & 66.0 & 0.81\\
    \cfg{bm25-lemma}          & 60.2 & 73.5 & 76.5 & 78.4 & 81.8 & 83.7 & 0.666 & 0.684 & 67.7 & 0.89\\
    \cfg{dense-surro-rrf}     & 54.9 & 75.4 & 79.5 & 83.0 & 86.0 & 87.5 & 0.643 & 0.673 & 66.7 & 0.84\\
    \cfg{dense}               & 52.3 & 68.2 & 71.6 & 72.7 & 74.6 & 74.6 & 0.590 & 0.616 & 61.7 & 0.63\\
    \DTF{}                    & 51.9 & 74.2 & 82.2 & \best{86.0} & \best{87.5} & \best{89.0} & 0.618 & 0.659 & \best{70.3} & \best{1.00}\\
    \cfg{bm25-expanded}       & 50.4 & 65.5 & 70.5 & 75.0 & 81.1 & 83.3 & 0.580 & 0.602 & 62.3 & 0.66\\
    \cfg{dense-rephrased}     & 50.0 & 69.3 & 73.5 & 76.5 & 78.0 & 78.0 & 0.581 & 0.614 & 65.0 & 0.77\\
    \cfg{dense-prf}           & 46.2 & 68.9 & 74.6 & 78.8 & 81.1 & 84.5 & 0.557 & 0.596 & 64.3 & 0.74\\
    \cfg{bm25-surrogate}      & 45.8 & 66.7 & 71.6 & 78.4 & 80.7 & 84.1 & 0.553 & 0.584 & 61.0 & 0.60\\
    \cfg{no-rerank}           & 44.7 & 61.4 & 68.6 & 75.0 & 80.3 & 83.0 & 0.524 & 0.555 & 56.7 & 0.41\\
    \cfg{dense-surro-rescore} & 41.3 & 58.7 & 67.0 & 72.7 & 74.6 & 74.6 & 0.490 & 0.526 & 58.7 & 0.50\\
    \midrule
    \multicolumn{11}{l}{\emph{Controls}}\\
    \cfg{oracle}              & 100.0 & 100.0 & 100.0 & 100.0 & 100.0 & 100.0 & 1.000 & 1.000 & 70.3 & 1.00\\
    \cfg{oracle-doc}          & \phantom{0}0.4 & \phantom{0}3.8 & \phantom{0}8.0 & 13.3 & 26.9 & 39.0 & 0.031 & 0.032 & 37.7 & $-0.40$\\
    \cfg{closed-book}         & \phantom{0}0.0 & \phantom{0}0.0 & \phantom{0}0.0 & \phantom{0}0.0 & \phantom{0}0.0 & \phantom{0}0.0 & 0.000 & 0.000 & 47.0 & 0.00\\
    \cfg{random}              & \phantom{0}0.0 & \phantom{0}0.0 & \phantom{0}0.0 & \phantom{0}0.0 & \phantom{0}0.0 & \phantom{0}0.4 & 0.000 & 0.000 & 25.3 & $-0.93$\\
    \bottomrule
  \end{tabular}
  \caption{Main results, pooled over the 2024 and 2025 examinations. Starred
    metrics are over the retrievable subset, $n^{*}=264$; every starred
    percentage is an integer count out of 264. Citation accuracy and $G$
    (Eq.~\ref{eq:gap}) are over all $n=300$. Ordered by Hit@1; note that
    ordering by Hit@20 or deeper places \DTF{} first, and that \DTF{} matches
    the oracle ceiling on citation accuracy. Document-level hit rate at
    $k{=}10$, cited in the text, is 92.0\% for \DTF{}, 90.5\% for \cfg{rrf},
    89.8\% for \ASCRH{} and 100\% for both oracles by construction. Three dense configurations show identical Hit@50 and Hit@100 because a
  depth-100 chunk list max-pools to fewer than 100 distinct articles.}
  \label{tab:main}
\end{table*}

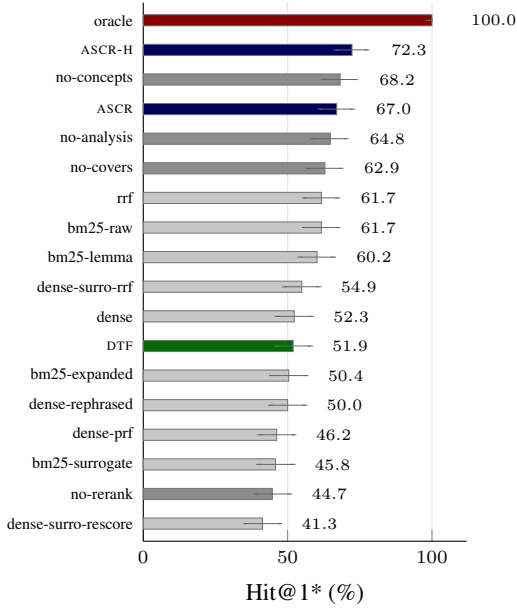
\begin{figure}[t]
  \centering
  \begin{tikzpicture}
    \pgfplotsset{
      barstyle/.style={
        xbar, bar shift=0pt, bar width=4.2pt, draw=black!55, line width=0.28pt,
        error bars/x dir=both, error bars/x explicit,
        error bars/error bar style={black!60, line width=0.35pt},
    error bars/error mark options={mark size=0.9pt, black!60}}}
    \begin{axis}[
        width=0.76\columnwidth, height=8.7cm,
        xmin=0, xmax=112,
        xlabel={Hit@1* (\%)}, xlabel near ticks,
        symbolic y coords={dense-surro-rescore, no-rerank, bm25-surrogate,
          dense-prf, dense-rephrased, bm25-expanded, triad, dense,
          dense-surro-rrf, bm25-lemma, bm25-raw, rrf, no-covers,
        no-analysis, full, no-concepts, full-hybrid, oracle},
        ytick={dense-surro-rescore, no-rerank, bm25-surrogate,
          dense-prf, dense-rephrased, bm25-expanded, triad, dense,
          dense-surro-rrf, bm25-lemma, bm25-raw, rrf, no-covers,
        no-analysis, full, no-concepts, full-hybrid, oracle},
        yticklabels={dense-surro-rescore, no-rerank, bm25-surrogate,
          dense-prf, dense-rephrased, bm25-expanded, {\textsc{dtf}}, dense,
          dense-surro-rrf, bm25-lemma, bm25-raw, rrf, no-covers,
        no-analysis, {\textsc{ascr}}, no-concepts, {\textsc{ascr-h}}, oracle},
        y tick label style={font=\tiny}, x tick label style={font=\scriptsize},
        label style={font=\small},
        nodes near coords, point meta=x,
        nodes near coords style={font=\tiny, text=black, xshift=11pt,
          /pgf/number format/fixed, /pgf/number format/zerofill,
        /pgf/number format/precision=1},
        every node near coord/.append style={anchor=west},
        axis x line*=bottom, axis y line*=left,
        xmajorgrids, grid style={gray!22, line width=0.3pt},
      enlarge y limits=0.035, clip=false]

      \addplot[barstyle, fill=black!22]
      table[x=v,y=m,x error plus=p,x error minus=n] {
        m                    v     p    n
        dense-surro-rescore 41.3  6.0  5.8
        bm25-surrogate      45.8  6.1  5.9
        dense-prf           46.2  6.0  5.9
        dense-rephrased     50.0  6.0  6.0
        bm25-expanded       50.4  6.0  6.0
        dense               52.3  5.9  6.0
        dense-surro-rrf     54.9  5.9  6.0
        bm25-lemma          60.2  5.7  6.0
        bm25-raw            61.7  5.7  5.9
        rrf                 61.7  5.7  5.9
      };
      \addplot[barstyle, fill=black!45]
      table[x=v,y=m,x error plus=p,x error minus=n] {
        m             v     p    n
        no-rerank    44.7  6.0  5.9
        no-covers    62.9  5.6  6.0
        no-analysis  64.8  5.5  6.0
        no-concepts  68.2  5.3  5.9
      };
      \addplot[barstyle, fill=blue!35!black]
      table[x=v,y=m,x error plus=p,x error minus=n] {
        m             v     p    n
        full         67.0  5.4  5.8
        full-hybrid  72.3  5.1  5.6
      };
      \addplot[barstyle, fill=green!42!black]
      table[x=v,y=m,x error plus=p,x error minus=n] {
        m       v     p    n
        triad  51.9  6.0  6.0
      };
      \addplot[barstyle, fill=red!52!black]
      table[x=v,y=m,x error plus=p,x error minus=n] {
        m        v      p    n
        oracle  100.0  0.0  1.4
      };
    \end{axis}
  \end{tikzpicture}
  \caption{Hit@1 with 95\% Wilson intervals, $n^{*}{=}264$. Proposed systems
    in blue (\ASCR{}, \ASCRH{}) and green (\DTF{}); \ASCRH{} ablations in dark
    grey; baselines in light grey; oracle ceiling in red. Intervals are marginal
    and therefore conservative: the paired tests in Table~\ref{tab:mcnemar}
  separate \ASCRH{} from every configuration shown except \cfg{no-concepts}.}
  \label{fig:recall1}
\end{figure}

\subsection{Rank one: \ASCRH{} leads every non-oracle configuration on point estimate}

\ASCRH{} places the reference provision first for 72.3\% of retrievable
questions (191 of 264), against 61.7\% for BM25, 61.7\% for reciprocal rank
fusion and 52.3\% for dense retrieval. Figure~\ref{fig:recall1} shows the full
ordering and Table~\ref{tab:mcnemar} the paired tests against every other
configuration and control. Of the twenty comparisons, eighteen are significant
in \ASCRH{}'s favour, sixteen of them at $p<0.001$; one is not significant; and
one, the oracle ceiling, is significant against it, as it must be. The
non-significant comparison is \cfg{no-concepts}, one of \ASCRH{}'s own
ablations, at $p=0.082$; the concept-matching term is therefore the weakest
component of the cascade at rank one, although Table~\ref{tab:main} shows it
costs 8.7 points by a cutoff of ten.

\begin{table}[t]
  \centering\small
  \setlength{\tabcolsep}{3.5pt}
  \begin{tabular}{lrrrrl}
    \toprule
    vs. & other & $\Delta$ & $b$ & $c$ & $p$\\
    \midrule
    \cfg{dense-surro-rescore} & 41.3 & $+31.0$ & 88 & \phantom{0}6 & $<$0.001\\
    \cfg{no-rerank}           & 44.7 & $+27.6$ & 76 & \phantom{0}3 & $<$0.001\\
    \cfg{bm25-surrogate}      & 45.8 & $+26.5$ & 78 & \phantom{0}8 & $<$0.001\\
    \cfg{dense-prf}           & 46.2 & $+26.1$ & 76 & \phantom{0}7 & $<$0.001\\
    \cfg{dense-rephrased}     & 50.0 & $+22.3$ & 66 & \phantom{0}7 & $<$0.001\\
    \cfg{bm25-expanded}       & 50.4 & $+21.9$ & 67 & \phantom{0}9 & $<$0.001\\
    \DTF{}                    & 51.9 & $+20.4$ & 66 & 12 & $<$0.001\\
    \cfg{dense}               & 52.3 & $+20.0$ & 63 & 10 & $<$0.001\\
    \cfg{dense-surro-rrf}     & 54.9 & $+17.4$ & 60 & 14 & $<$0.001\\
    \cfg{bm25-lemma}          & 60.2 & $+12.1$ & 47 & 15 & $<$0.001\\
    \cfg{bm25-raw}            & 61.7 & $+10.6$ & 44 & 16 & $<$0.001\\
    \cfg{rrf}                 & 61.7 & $+10.6$ & 44 & 16 & $<$0.001\\
    \cfg{no-covers}           & 62.9 & $+9.4$  & 29 & \phantom{0}4 & $<$0.001\\
    \cfg{no-analysis}         & 64.8 & $+7.5$  & 32 & 12 & 0.004\\
    \ASCR{}                   & 67.0 & $+5.3$  & 17 & \phantom{0}3 & 0.004\\
    \cfg{no-concepts}         & 68.2 & $+4.1$  & 22 & 11 & 0.082\\
    \midrule
    \cfg{closed-book}         & \phantom{0}0.0 & $+72.3$ & 191 & \phantom{0}0 & $<$0.001\\
    \cfg{random}              & \phantom{0}0.0 & $+72.3$ & 191 & \phantom{0}0 & $<$0.001\\
    \cfg{oracle-doc}          & \phantom{0}0.4 & $+71.9$ & 191 & \phantom{0}1 & $<$0.001\\
    \cfg{oracle}              & 100.0 & $-27.7$ & \phantom{0}0 & 73 & $<$0.001\\
    \bottomrule
  \end{tabular}
  \caption{Paired McNemar tests at rank one, baseline \ASCRH{} (72.3\%, 191 of
    264). $b$ and $c$ are discordant counts favouring \ASCRH{} and the
    comparison respectively; only these enter the test. Continuity-corrected
    \citep{edwards1948}, discordant cells computed within each examination year
  and summed. The final row is the oracle ceiling and is significant against \ASCRH{}.}
  \label{tab:mcnemar}
\end{table}

\subsection{Depth: the advantage does not persist}

\begin{figure}[t]
  \centering
  \begin{tikzpicture}
    \begin{axis}[
        width=0.97\columnwidth, height=5.6cm,
        xmode=log, log basis x=10,
        xtick={1,5,10,20,50,100}, xticklabels={1,5,10,20,50,100},
        xlabel={Cutoff $k$ (log scale)}, ylabel={Hit@$k$* (\%)},
        xmin=0.85, xmax=125, ymin=38, ymax=93,
        label style={font=\small}, tick label style={font=\scriptsize},
        grid=both, grid style={gray!20, line width=0.3pt},
        axis x line*=bottom, axis y line*=left,
        legend style={font=\tiny, draw=none, fill=none,
        at={(0.98,0.04)}, anchor=south east},
      legend cell align=left]
      \addplot[blue!55!black, line width=1.2pt, mark=*, mark size=1.6pt]
      coordinates {(1,72.3)(5,81.1)(10,83.7)(20,84.5)(50,86.0)(100,87.5)};
      \addlegendentry{\ASCRH{}}
      \addplot[green!45!black, line width=1.2pt, mark=triangle*, mark size=1.9pt]
      coordinates {(1,51.9)(5,74.2)(10,82.2)(20,86.0)(50,87.5)(100,89.0)};
      \addlegendentry{\DTF{}}
      \addplot[orange!80!black, mark=square*, mark size=1.3pt]
      coordinates {(1,61.7)(5,75.8)(10,80.3)(20,83.3)(50,85.6)(100,86.7)};
      \addlegendentry{rrf}
      \addplot[gray!55!black, dashed, mark=diamond*, mark size=1.4pt]
      coordinates {(1,61.7)(5,73.9)(10,76.1)(20,79.2)(50,83.0)(100,84.1)};
      \addlegendentry{bm25-raw}
      \addplot[gray!70!black, dotted, mark=o, mark size=1.3pt]
      coordinates {(1,52.3)(5,68.2)(10,71.6)(20,72.7)(50,74.6)(100,74.6)};
      \addlegendentry{dense}
    \end{axis}
  \end{tikzpicture}
  \caption{Hit rate against cutoff depth. \ASCRH{} leads through $k{=}10$ and
    \DTF{} from $k{=}20$, but the two are separated significantly only at
  $k{=}1$ and $k{=}5$; from $k{=}10$ onward the difference is within noise.}
  \label{fig:crossover}
\end{figure}
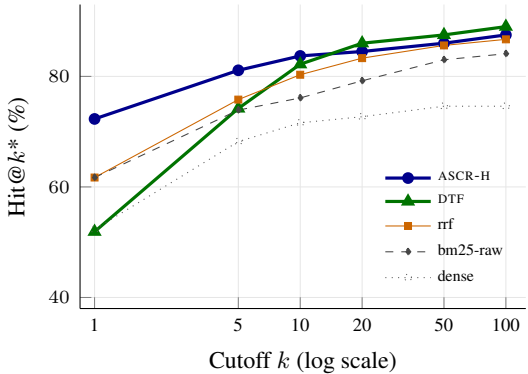

Figure~\ref{fig:crossover} plots Hit@$k$ for the two proposed systems and three
baselines. \ASCRH{} leads on point estimate at $k \in \{1,5,10\}$; at $k=20$
\DTF{} overtakes it (86.0 versus 84.5) and leads at every deeper cutoff,
reaching 89.0\% at $k=100$.

The paired tests qualify this picture in a way the point estimates alone do
not. \ASCRH{}'s advantage over \DTF{} is significant at $k=1$ ($b{=}66$,
$c{=}12$, $p<0.001$) and at $k=5$ ($b{=}29$, $c{=}11$, $p=0.007$), but not at
$k=10$ ($b{=}13$, $c{=}9$, $p=0.52$), and \DTF{}'s lead at $k=20$ is likewise
not significant ($b{=}6$, $c{=}10$, $p=0.45$). The honest reading is therefore
convergence rather than a crossover: reranking buys a large, reliable advantage
at the head, that advantage is gone by a cutoff of ten, and beyond it the two
designs are statistically indistinguishable on this sample even where their
point estimates diverge. \DTF{} also attains the highest document-level hit
rate of any non-oracle configuration (92.0\% at $k{=}10$) and the highest
citation accuracy of any configuration, matching the oracle at 70.3\%.

This is by design rather than by accident. \ASCRH{} spends two sequential model
calls to order a candidate set precisely; \DTF{} spends none, and instead
widens the candidate set through three complementary retrievers and a
deterministic prior. Precision at the head and coverage at depth are bought by
different mechanisms, and the mechanisms have very different costs
(Section~\ref{sec:cost}).

Two qualifications belong with this result. First, the generator in our
protocol reads only $k{=}10$ articles, so \DTF{}'s point-estimate leads at $k
\in \{20,50,100\}$ are properties of the retrieved list rather than of anything
the reader currently consumes; \citet{liu2024lost} report that reader
performance saturates well before retriever performance does. Second, \DTF{}'s
behaviour at depth is not separated from its act-scoping prior, which as
Section~\ref{sec:methods} shows partitions the candidate set rather than merely
tilting it, and which matched the correct act on every question in this
benchmark. We return to both in the Limitations section.

\subsection{Near-miss context is worse than none}

\cfg{oracle-doc}, which supplies the correct act but random articles within it,
reaches 100\% document-level hit rate by construction, 0.4\% Hit@1, and
citation accuracy of 37.7\%. That is 9.3 points \emph{below} the 47.0\% of a
control receiving no context at all. The drop is significant: 46 questions are
cited correctly by \cfg{closed-book} but not by \cfg{oracle-doc}, against 18 in
the other direction ($p<0.001$). \cfg{oracle-doc} also trails \ASCRH{} by 29.6
points ($b{=}98$, $c{=}9$, $p<0.001$). Plausible but incorrect provisions
displace correct
parametric knowledge, which suggests that act-level retrieval alone is not
merely insufficient but actively harmful.

The effect has precedent. \citet{liu2024lost} report that when the relevant document sits in the worst position within a twenty- or thirty-document context, GPT-3.5-Turbo falls below its own closed-book accuracy, and
\citet{legalragbench2026} find that retrieval quality is the primary driver of
end-to-end legal RAG performance and that most hallucinations in production
legal RAG systems are induced by retrieval failures. Our result isolates the same phenomenon
with the confound removed: \cfg{oracle-doc} holds the act correct and varies
only the article, so nothing about topical relevance explains the drop.

\section{Ablations and Negative Results}\label{sec:ablations}

\begin{figure}[t]
  \centering
  \begin{tikzpicture}
    \begin{axis}[
        xbar, bar shift=0pt, bar width=7pt,
        width=0.80\columnwidth, height=4.3cm,
        xmin=-36, xmax=2,
        xlabel={$\Delta$ Hit@1 when component removed (pp)},
        xlabel near ticks,
        symbolic y coords={reranker, covers, analysis, dense fusion, concepts},
        ytick={reranker, covers, analysis, dense fusion, concepts},
        y tick label style={font=\scriptsize},
        x tick label style={font=\scriptsize}, label style={font=\small},
        nodes near coords, point meta=x,
        nodes near coords style={font=\tiny, text=black, xshift= -4pt,
        /pgf/number format/fixed, /pgf/number format/precision=1},
        every node near coord/.append style={anchor=east},
        axis x line*=bottom, axis y line*=left,
        xmajorgrids, grid style={gray!22, line width=0.3pt},
      enlarge y limits=0.16, clip=false]
      \addplot[fill=black!25, draw=black!55, line width=0.28pt]
      coordinates {(-4.1,concepts) (-5.3,dense fusion) (-7.5,analysis)
      (-9.4,covers)};
      \addplot[fill=blue!35!black, draw=blue!25!black, line width=0.28pt]
      coordinates {(-27.6,reranker)};
    \end{axis}
  \end{tikzpicture}
  \caption{Component contributions to \ASCRH{}. Each bar is the change in
    Hit@1 when a single component is removed. Reranking accounts for nearly
    three times the next largest effect and is the only component with no cheaper
  substitute.}
  \label{fig:ablation}
\end{figure}
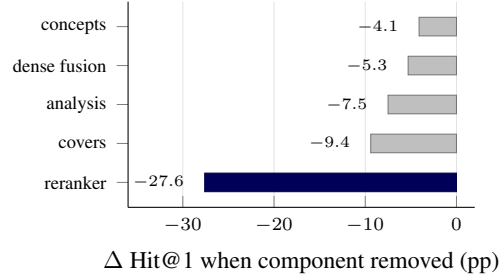

\subsection{Reranking is the dominant component}

\begin{table}[t]
  \centering\small
  \setlength{\tabcolsep}{4pt}
  \begin{tabular}{lrrrr}
    \toprule
    Removed & H@1* & H@10* & MRR* & $\Delta$H@1\\
    \midrule
    none (\ASCRH{})      & 72.3 & 83.7 & 0.764 & \\
    concepts             & 68.2 & 75.0 & 0.710 & $-4.1$\\
    query analysis       & 64.8 & 73.9 & 0.684 & $-7.5$\\
    covers ($Q_a$ term)  & 62.9 & 73.5 & 0.670 & $-9.4$\\
    dense fusion (=\ASCR{}) & 67.0 & 77.3 & 0.713 & $-5.3$\\
    reranker             & 44.7 & 68.6 & 0.524 & $\mathbf{-27.6}$\\
    \bottomrule
  \end{tabular}
  \caption{Component ablation on \ASCRH{}. Removing the reranker costs 27.6
  points of Hit@1, nearly three times the next largest effect.}
  \label{tab:ablation}
\end{table}

Removing the reranking stage reduces Hit@1 from 72.3\% to 44.7\% and MRR
from 0.764 to 0.524 ($b{=}76$, $c{=}3$, $p<0.001$). Hit@20 falls by only 9.5
points over the same ablation, confirming that reranking acts on ordering
rather than on candidate coverage. This is the largest effect in the study and
the only component with no cheaper substitute
(Figure~\ref{fig:ablation}).

\citet{dadas2024ranking} report that most Polish and multilingual rerankers underperform a strong dense retriever and generalise poorly, with the exception of models with a large number of parameters; their prescribed remedy is either domain fine-tuning or a larger reranker. Our reranker is \cfg{gemini-3.1-flash-lite}, the smallest tier of its family, so
on their result we should have seen little or no gain. Two things separate the
settings. Their rerankers are dedicated ranking models applied zero-shot to
open-domain passage retrieval, whereas ours is an instruction-following model
given a listwise prompt over forty candidates; and they themselves distil a 435M
reranker that surpasses their 13B teacher, so parameter count is not the
operative variable even within their own study.

A task-structure difference we think amplifies the effect is visible in PIRB
\citep{dadas2024pirb}: averaged over its 41 tasks, BM25 reaches 41.9 nDCG@10
against 57.3 for \texttt{multilingual-e5-large}, a lexical-to-dense gap of
roughly fifteen points in the dense model's favour. On our task the ordering reverses: BM25
reaches 61.7\% Hit@1 against 52.3\% for dense. The reversal is not peculiar to
our benchmark: \citet{rybak2024silver} report that on Polish legal passage
retrieval BM25 outperforms every neural model they test, and attribute this to
passage length and to high lexical overlap between questions and passages. With
exactly one correct
article among many lexically similar candidates drawn from the same act,
ordering at the head carries most of the available signal, and lexical evidence
is unusually informative. Under these conditions a capable reranker has more to
work with than in open-domain passage retrieval, where several retrieved
passages may be acceptable.

\subsection{Generated representations help only in fusion}

The three BM25 variants separate where generation is applied. At Hit@1,
\cfg{bm25-raw} (no generation) scores 61.7; \cfg{bm25-surrogate} (index-side
only) 45.8; \cfg{bm25-expanded} (query-side only) 50.4. Neither generated
representation beats verbatim text on its own, and index-side is the worse of
the two standalone.

The result reverses under fusion. Inside \ASCRH{} the surrogate index
contributes materially: the $Q_a$ term is worth 9.4 points of Hit@1
(Table~\ref{tab:ablation}), and removing it costs more than removing query
analysis. \DTF{} also draws on the surrogate index through
$R_{\mathrm{cov}}$, but we report no ablation of that branch, so its
contribution there is untested. Generated representations are therefore useful as a
\emph{complementary} signal but not as a replacement for the statutory text,
which is consistent with their function: they close the vocabulary gap on
questions whose phrasing does not match the statute, and add noise on questions
whose phrasing does.

One caveat bounds the standalone result. \cfg{bm25-surrogate} substitutes the
surrogate fields for $\xi(a)$, whereas document expansion as introduced by
\citet{nogueira2019doc2query} appends predicted queries to the document and
keeps it, reporting a gain of roughly 15\% from doing so. Our $-15.9$-point
standalone result is therefore evidence about substitution, not about generated
representations as such; the concatenated index is the configuration that would
settle it, and we did not run it (see Limitations).

\subsection{Three negative results}

\paragraph{Lemmatisation does not help.}
\cfg{bm25-lemma} scores 60.2\% Hit@1 against 61.7\% for \cfg{bm25-raw}, and
trails at every cutoff except $k{=}10$, where it leads by 0.4 points. Despite
seven-case inflection, suffix stripping yields no benefit, because examination
questions quote statutory language closely and therefore already share surface
forms with the target. This sits in mild tension with PIRB
\citep{dadas2024pirb}, whose BM25 baseline lemmatises by default through
Morfologik; the difference is most likely the closeness of our questions to the
statutory text, and it may not transfer to free-form practitioner queries.

\paragraph{Pseudo-relevance feedback harms head precision.}
\cfg{dense-prf} scores 46.2\% Hit@1 against 52.3\% for \cfg{dense}, a loss of
6.1 points, while gaining 6.1 points at $k{=}20$ and 9.9 at $k{=}100$.
Expansion widens the candidate pool at the cost of disordering its head. This
is the classic pseudo-relevance feedback trade-off, and on a task where only
rank one matters it is unfavourable.

\paragraph{Query rewriting does not help.}
\cfg{dense-rephrased} scores 50.0\% against 52.3\% for \cfg{dense}, and
\cfg{bm25-expanded} 50.4\% against 61.7\% for \cfg{bm25-raw}. Both incur an
additional model call and roughly 1.1\,s of latency for no measurable gain.
\cfg{no-analysis}, which removes the equivalent stage from \ASCRH{}, costs 7.5
points of Hit@1 while saving \$0.036 per hundred queries, so query-side
generation is useful inside the cascade, where it conditions the fields the
article stage and the reranker see, but not as a standalone modification to the
query.

\subsection{Ranking gains do not transfer to citation accuracy}

\ASCRH{} leads every ranking metric yet reaches citation accuracy of 67.3\%
(202 of 300), statistically indistinguishable from \cfg{bm25-raw} (66.0\%,
$b{=}19$, $c{=}15$, $p{=}0.61$), \cfg{rrf} (67.7\%, $b{=}15$, $c{=}16$,
$p{=}1.00$), \cfg{bm25-lemma} (67.7\%, $b{=}17$, $c{=}18$, $p{=}1.00$) and
\DTF{} (70.3\%, $b{=}14$, $c{=}23$, $p{=}0.19$), which scores numerically
higher and matches the oracle. Twelve of the twenty comparisons are
significant: four are \ASCRH{}'s own ablations, three are controls, and the
remaining five are \ASCR{} and four configurations that trail \ASCRH{} by at
least twenty points at rank one. The eight that are not significant include
every strong single-call baseline and the oracle itself.

Placing the reference provision at rank one rather than rank five therefore
does not change how often the generator names it. What does change citation
accuracy is whether the provision reaches the context window at all: every
retrieval configuration scores above the closed-book floor of 47.0\% and
\cfg{oracle-doc} falls below it, and the one paired comparison we test,
\ASCRH{} against \cfg{closed-book}, is significant ($b{=}83$, $c{=}22$, $p<0.001$). We read this as evidence that on this task
the generator is tolerant of position within the retrieved context but
sensitive to presence.

The clearest precedent is the ceiling argument of
\citet{reasoningfocused2025}, who report that improvements in Recall@10 do not
reliably translate into downstream question-answering gains because the gain is
bounded by how much the reader can extract from the gold passage at all. Our
$G$ metric measures exactly that bound: the achievable band between a 47.0\%
closed-book floor and a 70.3\% oracle ceiling is 23.3 points, and no ranking
improvement can produce more than that. \DTF{} already closes all of it, with a
retrieval system that trails \ASCRH{} by twenty points at rank one.
\citet{legalragbench2026} observe the presence-driven form of the same effect,
identifying retrieval quality rather than generation as the primary driver of
end-to-end performance.

Our result is more circumscribed than the position effect of
\citet{liu2024lost}, and we do not claim to contradict them. Their
ten-document condition is our $k{=}10$, so context length is not the
difference. What differs is how much the reader depends on the context at all:
in their multi-document setting the gap between closed-book and oracle answer
accuracy ranges from 28 points for Claude-1.3 to 50 points for MPT-30B-Instruct,
whereas in ours it is 1.6 points on answer accuracy and 23.3 on citation
accuracy, and a generator that answers correctly
without the context has little room to display sensitivity to position within
it. Their dependent variable is also answer accuracy, ours citation accuracy.
The practical implication for this task is that effort spent moving the correct
provision from rank five to rank one is not repaid downstream, and that
rank-one accuracy should not be treated as a proxy for answer quality.

\section{Cost and Latency}\label{sec:cost}

\begin{table}[t]
  \centering\small
  \setlength{\tabcolsep}{4pt}
  \begin{tabular}{lrrrrr}
    \toprule
    Configuration & H@1* & H@20* & p50 & p95 & USD\\
    & (\%) & (\%) & (ms) & (ms) & /100q\\
    \midrule
    \ASCRH{}   & \best{72.3} & 84.5 & 7784 & 13990 & 0.230\\
    \ASCR{}    & 67.0 & 78.0 & 7997 & 13910 & 0.233\\
    \midrule
    \cfg{no-analysis} & 64.8 & 74.2 & 9096 & 13284 & 0.194\\
    \cfg{no-rerank} & 44.7 & 75.0 & 7035 & 13283 & 0.145\\
    \cfg{bm25-expanded} & 50.4 & 75.0 & 1856 & \phantom{0}2231 & 0.165\\
    \cfg{dense-rephrased} & 50.0 & 76.5 & 1842 & \phantom{0}2196 & 0.156\\
    \midrule
    \cfg{rrf}  & 61.7 & 83.3 & \phantom{0}830 & \phantom{0}1188 & 0.114\\
    \cfg{bm25-raw} & 61.7 & 79.2 & \phantom{0}726 & \phantom{0}1018 & 0.118\\
    \DTF{}     & 51.9 & \best{86.0} & \phantom{0}820 & \phantom{0}1090 & 0.108\\
    \cfg{dense} & 52.3 & 72.7 & \phantom{0}733 & \phantom{0}1093 & 0.108\\
    \midrule
    \cfg{closed-book} & \phantom{0}0.0 & \phantom{0}0.0 & \phantom{0}668 & \phantom{00}885 & 0.014\\
    \bottomrule
  \end{tabular}
  \caption{Cost and latency. Horizontal rules group configurations by total
    model calls including generation: \ASCRH{} and \ASCR{} make three,
    \cfg{no-analysis} through \cfg{dense-rephrased} make two, and the rest make
    one. Generation cost is identical across configurations, so all
  differences arise from pre-generation calls, each a sequential round trip.}
  \label{tab:cost}
\end{table}

Latency is bimodal. Counting the generation call, the single-call configurations of Table~\ref{tab:cost}
complete in 726--830\,ms, and 668\,ms for the closed-book control, which
retrieves nothing; three-call configurations take 7.8--8.0\,s. There is no
intermediate operating point
among the strong configurations, because each pre-generation call is a
sequential round trip rather than a parallelisable computation. The \ASCRH{}
dense branch is the exception that proves the rule: it runs concurrently with
query analysis and therefore costs no wall-clock time at all, which is why \ASCRH{} costs no more wall-clock time than \ASCR{} despite retrieving more.

The two-call band spans 1.8 to 9.1\,s because the calls are not equivalent. The
query-analysis call retained by \cfg{no-rerank} and the reranking call retained
by \cfg{no-analysis} are both long structured prompts over a hundred
candidates; \cfg{bm25-expanded} and \cfg{dense-rephrased} issue a one-line
rewrite instead and run under 2\,s. Removing a call therefore does not
straightforwardly reduce latency: \cfg{no-analysis} is slower than full
\ASCRH{}, because the reranker it retains sees a worse-ordered candidate list
and returns longer output. Cost, which tracks tokens rather than round trips,
does fall as expected.

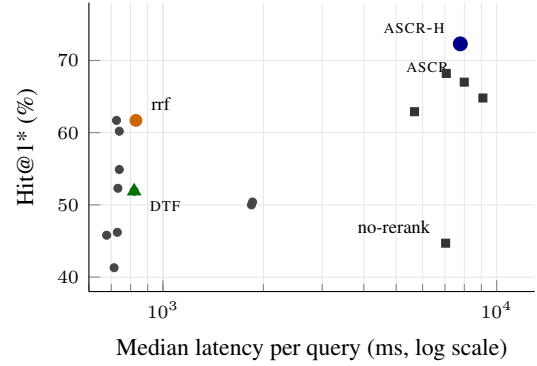
\begin{figure}[t]
  \centering
  \begin{tikzpicture}
    \begin{axis}[
        width=0.97\columnwidth, height=5.4cm,
        xmode=log, log basis x=10,
        xlabel={Median latency per query (ms, log scale)},
        ylabel={Hit@1* (\%)},
        xmin=600, xmax=13000, ymin=38, ymax=78,
        label style={font=\small}, tick label style={font=\scriptsize},
        grid=both, grid style={gray!20, line width=0.3pt},
      axis x line*=bottom, axis y line*=left, clip=false]
      coordinates {(678,45.8)(726,61.7)(7784,72.3)};
      \addplot[only marks, mark=*, mark size=1.5pt, color=gray!55!black]
      coordinates {(726,61.7)(740,60.2)(733,52.3)(820,51.9)(740,54.9)
      (730,46.2)(678,45.8)(714,41.3)(1856,50.4)(1842,50.0)};
      \addplot[only marks, mark=square*, mark size=1.5pt, color=gray!40!black]
      coordinates {(7997,67.0)(7035,44.7)(9096,64.8)(7064,68.2)(5674,62.9)};
      \addplot[only marks, mark=*, mark size=2.6pt, color=blue!55!black]
      coordinates {(7784,72.3)};
      \addplot[only marks, mark=triangle*, mark size=2.8pt, color=green!45!black]
      coordinates {(820,51.9)};
      \addplot[only marks, mark=*, mark size=2.2pt, color=orange!80!black]
      coordinates {(830,61.7)};
      \node[font=\scriptsize, anchor=south east] at (axis cs:7784,72.3) {\ASCRH{}\;};
      \node[font=\scriptsize, anchor=south east] at (axis cs:7997,67.0) {\ASCR{}\;};
      \node[font=\scriptsize, anchor=south west] at (axis cs:830,61.7) {\;rrf};
      \node[font=\scriptsize, anchor=north west] at (axis cs:820,51.9) {\;\DTF{}};
      \node[font=\scriptsize, anchor=south east] at (axis cs:7035,44.7) {no-rerank\;};
    \end{axis}
  \end{tikzpicture}
  \caption{Cost and quality at rank one. Median latency against Hit@1* for every
    configuration in Table~\ref{tab:cost} together with several further
    baselines. The interval between \cfg{bm25-raw} and \ASCRH{} is bought by two
  additional sequential model calls.}
  \label{fig:pareto}
\end{figure}

Whether the head-precision premium is worthwhile is a deployment question. For
interactive search a 14-second 95th-percentile latency is disqualifying; for
asynchronous review or batch enrichment it is immaterial, and 10.6 points of
rank-one accuracy over the best call-free configuration is substantial. It is
also cheap relative to the reranking literature: \ASCRH{} costs \$0.0023 per
query end to end, against the \$0.596 per query \citet{sun2023chatgpt} report
for listwise GPT-4 reranking of a hundred passages, roughly two orders of
magnitude less, a factor of 259, at a median 7.8\,s against the 32\,s they report for GPT-4. A single
listwise call over forty candidates, rather than ten sliding-window calls over a
hundred passages, accounts for most of that difference.

\DTF{} occupies the opposite corner. Replacing the surrogate cascade and
reranker with three retrievers and a deterministic prior reduces latency to
820\,ms and cost to \$0.108 per hundred queries, a factor of 9.5 and 2.1
respectively, while attaining the highest Hit@20 (86.0\%), Hit@100 (89.0\%),
document hit rate (92.0\%) and citation accuracy (70.3\%, level with the
oracle) of any non-oracle configuration. It forfeits 20.4 points at rank one.
Cheaper components preserve coverage and surrender ordering, which is the trade
the design was built to make. On the metric that reflects what the generator
produces, the surrender costs nothing.
\citet{llmretrievercost2026} construct the analogous frontier for retrievers,
plotting quality against throughput and identifying non-dominated models; they
also caution that zero-cost lexical fusion, in their case linear, reliably helps weak retrievers but can harm strong ones, which bounds how far \DTF{}'s design would generalise to
a setting with a stronger base encoder.

\section{Conclusion}

We presented three surrogate-based retrieval designs for Polish statutory law
and evaluated them against fourteen baselines and four controls on 300
ministry-published examination questions. \ASCRH{} places the reference
provision at rank one significantly more often than every other non-oracle configuration
tested except one of its own ablations, and \DTF{} matches it from a cutoff of
ten onward at one
ninth the latency while closing the entire achievable band on citation
accuracy. The two occupy opposite ends of a cost--quality frontier and are
intended to be chosen between rather than ranked. Reranking accounts for 27.6
points of rank-one accuracy and is the only component without a cheaper
substitute; lemmatisation, pseudo-relevance feedback and query rewriting each
fail to improve retrieval on this task. The most consequential negative result
is that none of the rank-one advantage reaches the generated answer.

\section*{Limitations}

\paragraph{Surrogate coverage is correlated with the benchmark.}
Surrogate annotations cover 22{,}241 of 82{,}508 articles (27.0\%) but all 264
retrievable reference provisions. Annotation was prioritised by act importance,
and examination questions draw disproportionately on the same major codes and
the Constitution, so the two selections are correlated rather than independent.
Every configuration that matches against surrogate fields, which includes \ASCR{}, \ASCRH{} and its four ablations, \cfg{bm25-surrogate}, \cfg{dense-surro-rrf}, \cfg{dense-surro-rescore}, \cfg{dense-prf} and the $R_{\mathrm{cov}}$ branch of \DTF{}, therefore scores a fully
annotated target against a candidate field that is largely unannotated, and
some unknown share of their rank-one advantage is attributable to that
asymmetry rather than to the retrieval design. We did not measure the surrogate
coverage of the non-gold candidates that those systems actually rank, which is
the measurement that would bound the effect, and we did not repeat the
comparison on a coverage-matched subset. Both are the first things we would add.
The confound does not affect the configurations that never touch surrogate
fields (\cfg{bm25-raw}, \cfg{bm25-lemma}, \cfg{bm25-expanded}, \cfg{dense},
\cfg{dense-rephrased}, \cfg{rrf} and the controls), so the reversals among those, lexical over dense in particular, are
unaffected. Extending coverage to the full corpus would likely change the
balance between \ASCRH{} and \DTF{}, in an untested direction.

\paragraph{\DTF{}'s act-scoping prior is not isolated.}
As Section~\ref{sec:methods} sets out, $\beta$ sits above the attainable range
of the fused score, so act scoping partitions the candidate set for non-repealed articles rather than
tilting it. Every question in this benchmark names its act, and the matcher
identified the correct one in all 300 cases. We do not report a $\beta{=}0$
ablation, so we cannot say how much of \DTF{}'s behaviour from $k{=}10$ onward
is fusion and how much is the partition. Those results therefore hold for \DTF{} \emph{on questions that name their governing act}, and we do not claim them for tri-signal fusion in general.

\paragraph{Statistical power at depth.}
The comparisons that matter most at depth rest on few discordant pairs:
\ASCRH{} against \DTF{} produces 22 discordant questions at $k{=}10$ and 16 at
$k{=}20$, out of 264. We report these as null results rather than as evidence
of equivalence; a larger sample could separate the two designs in either
direction.

\paragraph{Question format.}
Examination items are multiple-choice, quote statutory language closely, and
name their governing act explicitly in all 300 cases. All three properties make
the task easier than practitioner queries, and the third is exploited directly
by \DTF{}'s act-scoping prior.

\citet{smywinski2025statement} make this point explicitly: on Polish legal
retrieval they observe a gap of roughly forty-five points of nDCG@10 between
datasets whose questions are derived from the target provisions and datasets of
questions posed by laypeople, and warn that the former yield performance
estimates that are too optimistic. Our questions are of the former kind.
The close lexical overlap also biases the comparison between retrieval families. Questions that quote statutory wording favour term matching directly, which is a likely explanation for BM25 outscoring dense retrieval here when the ordering is reversed on open-domain Polish benchmarks \citep{dadas2024pirb}. \citet{rybak2024silver} report the same reversal on Polish legal retrieval, attributing it in part to high lexical overlap between questions and passages and in part to passage length. We read this as evidence that the overlap is common across Polish legal retrieval resources rather than as evidence that it is benign. Absolute figures reported here should be read as an upper bound on what the same systems would achieve on practitioner queries; the comparisons between configurations, which is what the paper argues about, are affected only where a configuration exploits the format directly.

\paragraph{Possible training-data contamination.}
The entrance examinations are published by the Ministry of Justice together
with their answer keys, are widely reproduced, and supply the training questions
of at least one existing Polish legal dataset. Our closed-book control answers
92.7\% of items correctly, against accuracies in the mid-60s to high-70s reported by \citet{karp2026judge} for comparable models on a comparable Polish legal examination. A stronger generator, an easier examination and contamination are
all consistent with that gap, and we cannot distinguish them; we did not
pre-train or filter to exclude the test material, as \citet{nogueira2019passage}
do for TREC-CAR. Contamination would inflate answer accuracy and, to a lesser
degree, citation accuracy, and is a further reason not to read answer accuracy
as a retrieval metric. It does not threaten the between-configuration
comparisons, since the same generator is held fixed across all of them.

\paragraph{Article-level citation matching.}
Citation accuracy is scored at article level, but 69\% of reference provisions
specify a § or \emph{ust.} subdivision that is not scored. Reported figures are
therefore an upper bound on practitioner-grade precision. The automatic matcher
was not validated against human judgement, and
\citet{multilegalbench2026} report for the analogous norm-extraction task that
measured performance is dominated by the evaluation methodology rather than by
model capability.

\paragraph{Coverage ceiling.}
Nine of 300 reference citations could not be resolved to any article by our
parser, and 27 more resolve to articles absent from the corpus.
Starred metrics exclude all 36, but the exclusion is not random: missing
provisions skew toward less frequently examined acts. The same skew is
documented for BSARD, where only 1{,}612 of 22{,}633 articles are ever cited as
relevant and roughly 80\% of those come from four codes
\citep{louis2022statutory}.

\paragraph{An untested baseline.}
\cfg{bm25-surrogate} replaces the statutory text with the surrogate fields,
whereas document expansion in the sense of \citet{nogueira2019doc2query}
appends them to it. We do not report BM25 over $\xi(a)\Vert Q_a$, so our
standalone negative result on index-side generation is evidence about
substitution rather than about expansion.

\paragraph{Single generator, encoder and run.}
All results use \cfg{gemini-3.1-flash-lite} at temperature zero for surrogate
generation, query analysis, reranking and answer generation, and
\texttt{multilingual-e5-large} as the encoder. Each configuration was run once.
We therefore report no run-to-run variance, and the Wilson intervals capture
sampling variation over questions only. All stages run at temperature zero, but
we did not verify that repeated runs of the reranking stage return identical
permutations, so the intervals may be optimistic in that respect. The saturation of answer accuracy in particular
depends on the generator's parametric knowledge of Polish law and may not
transfer to weaker models. Newer general-purpose encoders have since been reported stronger than
\texttt{multilingual-e5-large} on Polish legal retrieval
\citep{smywinski2025statement}, so our dense branch is a conservative
instantiation rather than a ceiling.

\paragraph{Reference multiplicity.}
Six of the 300 questions carry more than one reference article, five naming two
and one naming three, and Hit@$k$ credits any of them. We do not report per-provision recall, so the paper does
not measure whether a system retrieves the complete set of governing provisions
where more than one applies. Retrieval of dependent provisions, cross-references
and delegated regulations is likewise not measured, although these determine
whether a retrieved provision can be applied correctly.

\paragraph{Temporal correctness untested.}
Each examination is fixed to a published legal state, which would permit
evaluating whether a system retrieves the version of a provision in force at
that date. We do not exploit this; it is the most immediate extension of the
benchmark.

\paragraph{Hyperparameters.}
Fusion weights, $k_0$, the BM25 parameters and the re-scoring coefficients in
Eq.~\ref{eq:rescore} were set by hand and not tuned on a development split. They
are therefore neither optimised nor contaminated by the test set. Two
qualifications: $\beta$ and $\pi$ were chosen above the attainable range of the
fused score, which is a design decision rather than an arbitrary value and is
described as such in Section~\ref{sec:methods}; and $k_1{=}1.2$, $b{=}0.75$ are
the common defaults, which \citet{robertson2009bm25} explicitly decline to
recommend as universal, while \citet{louis2022statutory} tune to $k_1{=}1.0$,
$b{=}0.6$ for statutory articles. BM25 is our strongest single baseline and a
\DTF{} component, so untuned defaults on a length-atypical corpus may
understate it.

\paragraph{Reader budget.}
The generator reads $k{=}10$ articles, so differences between systems at deeper
cutoffs are not visible to it in this protocol. Those cutoffs measure the
quality of the retrieved list, which is what this paper studies, and they bear
on settings with a larger context budget or a human reader; no result here
shows that they improve generated answers.

\paragraph{Sample size.}
Paired tests draw on 264 retrievable questions across two examination years, or
300 for citation and answer accuracy. Comparisons with fewer than roughly
twenty discordant pairs have limited power irrespective of the difference in
point estimates; discordant counts are reported throughout so that this can be
assessed per comparison.

\section*{Ethics Statement}

The benchmark is built from examination materials published by the Polish
Ministry of Justice together with their official answer keys, and from
statutory texts that are public by law. No personal data is involved and no
human annotation was commissioned.

The systems described here retrieve provisions rather than interpret or apply
them, and none of our metrics measures legal correctness. Our own results argue
against deployment as a source of legal answers: citation accuracy peaks at
70.3\% even with the reference provision in context, and article-level scoring
ignores the subdivision that 69\% of reference provisions specify. Any use in
practice belongs behind qualified human review.

\bibliography{custom}
\end{document}